\documentclass{article}
\usepackage{spconf,amsmath,graphicx}
\usepackage{siunitx}
\usepackage{tabularx,booktabs}
\usepackage{gensymb}
\usepackage{titlesec}
\usepackage{cite}
\usepackage{amsfonts}
\usepackage[dvipsnames]{xcolor}
\usepackage{hyperref}
\hypersetup{colorlinks=true,linkcolor=blue,citecolor=black}
\usepackage{caption}
\usepackage{pgfplots}
\pgfplotsset{every tick label/.append style={font=\footnotesize}}
\pgfplotsset{
  /pgfplots/my xbar legend/.style={
    /pgfplots/legend image code/.code={%
      \draw [#1] (-0.1cm,-0.1cm) rectangle (0.17cm,0.17cm);}
}}
\usepackage{fancyhdr}

\fancypagestyle{FirstPage}{
\lfoot{\copyright 2022 IEEE. Personal use of this material is permitted. Permission from IEEE must be obtained for all other uses, in any current or future media, including reprinting/republishing this material for advertising or promotional purposes, creating new collective works, for resale or redistribution to servers or lists, or reuse of any copyrighted component of this work in other works.}
}

\title{L2CS-Net: Fine-Grained Gaze Estimation in Unconstrained Environments}
\name{Ahmed A.Abdelrahman, Thorsten Hempel, Aly Khalifa, Ayoub Al-Hamadi\thanks{This research was funded by the Federal Ministry of Education and Research of Germany (BMBF) RoboAssist~no.~03ZZ0448L.}}
\address{Faculty of Electrical Engineering and Information Technology, Neuro-Information Technology Group\\
Otto-von-Guericke-University, Magdeburg, Germany}
\begin{document}
%
\maketitle
\begin{abstract}
Human gaze is a crucial cue used in various applications such as human-robot interaction and virtual reality. Recently, convolution neural network (CNN) approaches have made notable progress in predicting gaze direction. However, estimating gaze in-the-wild is still a challenging problem due to the uniqueness of eye appearance, lightning conditions, and the diversity of head pose and gaze directions. In this paper, we propose a robust CNN-based model for predicting gaze in unconstrained settings. We propose to regress each gaze angle separately to improve the per-angel prediction accuracy, which will enhance the overall gaze performance. In addition, we use two identical losses, one for each angle, to improve network learning and increase its generalization. We evaluate our model with two popular datasets collected with unconstrained settings. Our proposed model achieves state-of-the-art accuracy of 3.92\degree and 10.41\degree on MPIIGaze and Gaze360 datasets, respectively. We make our code open source at \url{https://github.com/Ahmednull/L2CS-Net}.
\end{abstract}
\section{Introduction}
\label{sec:intro}
\thispagestyle{FirstPage}
Eye gaze is one of the essential cues used in a large variety of applications. It indicates the user's level of engagement in human-robot interaction \cite{hempel2020slam,strazdas2022robot}, and open dialogue systems \cite{li2012vision}. Furthermore, it is used in augmented reality \cite{patney2016towards} to predict the users' attention, which improves the device's awareness and reduces power consumption. Therefore, researchers developed multiple methods and techniques for accurately estimating the human gaze. These methods are divided into two categories: model-based and appearance-based approaches. Model-based methods generally require dedicated hardware that makes them difficult to use in an unconstrained environment. On the other hand, appearance-based methods regress the human gaze directly from the images captured by inexpensive off-the-shelf cameras, making them easy to generate in different locations with unconstrained settings.

Recently, CNN-based appearance-based methods are the most commonly used gaze estimation methods as they provide better gaze performance \cite{cvpr2021,ca-net, dilated,liu2020gaze}. Most of the related work \cite{Gaze360,Mpiigaze,cvpr2021,dilated,ca-net} focussed on developing novel CNN-based networks which mainly consist of popular backbones (e.g. VGG \cite{rtgene}, ResNet-18 \cite{Gaze360}, ResNet-50 \cite{Ethxgaze}) to extract gaze features and finally outputs the gaze direction. The input to these networks can be a single stream \cite{Ethxgaze, Gaze360}(e.g., face or eye images) or multiple streams \cite{cvpr2021} (e.g., face and eye images). The most common loss function used for the gaze estimation task is the mean square loss or $\ell_2$ loss. However, Petr et al. \cite{Gaze360} proposed a novel pinball loss that estimates the gaze direction and error bounds together, which improves the accuracy of gaze estimation, especially in unconstrained settings. Although CNN-based methods achieve improved gaze accuracy, they lack robustness and generalization, especially in unconstrained environments.

In this paper, we introduce a new method to estimate 3D gaze angles from RGB images using a multi-loss approach. We propose to regress each gaze angle (yaw, pitch) independently using two fully-connected layers to enhance the prediction accuracy of each angle. Furthermore, we use two separate loss functions for each gaze angle. Each loss consists of gaze binary classification and regression components. Finally, the two losses are backpropagated through the network, which accurately fine-tunes the network weights and increases network generalization. We perform gaze bin classification by utilizing a softmax layer along with cross-entropy loss so that the network estimates the neighborhood of the gaze angle in a robust manner. Based on the proposed loss function and the softmax layer ($\ell_2$ loss+ cross-entropy loss+ softmax layer), we present a new network (L2CS-Net) to predict 3D gaze vector in unconstrained settings. Finally, we evaluate the robustness of our network on two popular datasets, MPIIGaze \cite{Mpiigaze} and Gaze360 \cite{Gaze360}. The proposed L2CS-Net achieved state-of-the-art performance on MPIIGaze and Gaze360 datasets.


\section{Related Work}
\label{sec:relatedwork}
According to the literature, appearance-based gaze estimation can be divided into conventional and CNN-based methods. Conventional gaze estimation methods use a regression function to create a person-specific mapping function to the human gaze, e.g., adaptive linear regression \cite{lu2014adaptive} and gaussian process regression \cite{williams2006sparse}. These methods show reasonable accuracy in constrained setup (e.g., subject-specific and fixed head pose and illumination), however they significantly decrease when tested on unconstrained settings. 

Recently, researchers have gained more interest in CNN-based gaze estimation methods, as they can model a highly nonlinear mapping function between images and gaze. Zhang et al. \cite{Mpiigaze} first proposed a simple VGG CNN-based architecture to predict gaze using a single eye image. Also, they designed a spatial weights CNN in \cite{Mpiigaze1} to give more weight to those regions of the face that related to the gaze in appearance. Krafka et al. \cite{itracker} proposed a multichannel network that takes eye images, full-face images, and face grid information as inputs.

Combining statistical models with deep learning is a good solution for gaze estimation as in \cite{MeNets}, which they introduced a mixed effect model that integrates information from statistics within CNN architecture based on eye images. Chen et al. \cite{dilated} adopted dilated convolutions to make use of the high-level features extracted from images without decreasing spatial resolution. In addition, they expanded their work by proposing GEDDNet \cite{GEDDNet}, which utilizes both gaze decomposition with dilated convolutions and reported better performance than using dilated convolutions only. Fischer et al. \cite{rtgene} append the head pose vector along with features extracted using a VGG CNN with eye crops to predict gaze angels. Additionally, they used an ensemble scheme to improve gaze accuracy.

Motivated by the two-eye asymmetry property, Cheng et al. \cite{farenet} proposed FAR-Net that estimates 3D gaze angels for both eyes with an asymmetric approach. They give asymmetric weights to each loss of the two eyes and finally sum these losses. The proposed model presented a top performance in multiple public datasets. Wang et al. \cite{BayesianApproach} integrated adversarial learning with the Bayesian approach in one framework, which demonstrates an increased gaze generalization performance. Cheng et al. \cite{ca-net} proposed a coarse-to-fine adaptive network (CA-Net) that first uses face image to predict primary gaze angels and adapt it with the residual estimated from eye crops. Then, they proposed a bi-gram model to bridge the primary gaze with the eye residual. Kellnhofer et al. \cite{Gaze360} used a temporal model (LSTM) with a sequence of 7 frames to predict gaze angels. In addition, they adopt pinball loss to jointly regress the gaze direction and error bounds together to improve gaze accuracy.

The most recent work with gaze estimation is AGE-Net \cite{cvpr2021} which they propose two parallel networks for each eye image, one is used to generate a feature vector using CNN, and the other is used to generate a weight feature vector using an attention-based network. The output of the two parallel networks is multiplied and then refined with the output of VGG CNN of the face images.
\begin{figure*}
  \includegraphics[width=\textwidth]{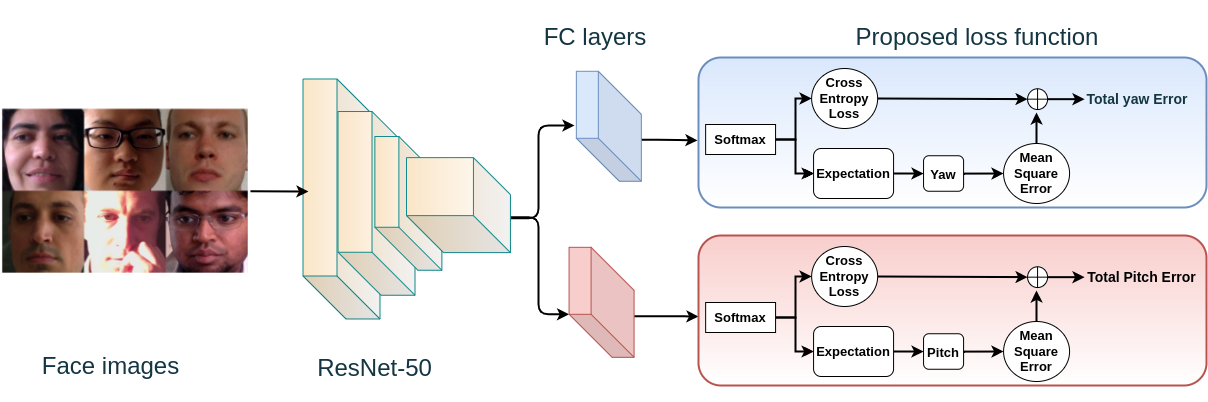}
  \caption{L2CS-Net with combined classification and regression losses.}
\label{fig:concept}
\end{figure*}

\section{METHOD}
\label{sec:method}
\subsection{Proposed loss function}
Most CNN-based gaze estimation models predict 3D gaze as the gaze direction angles (yaw, pitch) in spherical coordinates. Further, they adopt the mean-squared error ($\ell_2$ loss) for penalizing their networks. We propose to use two identical losses for each gaze angle. Each loss contains a combined cross-entropy loss and mean-squared error. Instead of directly predicting continuous gaze angels, we used a softmax layer with cross-entropy to predict binned gaze classification. Then, we estimate the expectation of the gaze binned output to fine-grain the predictions. Finally, we add a mean-squared error to the output to improve the gaze predictions. Using $\ell_2$ together with Softmax can tune the nonlinear softmax layer with immense flexibility.

The cross entropy loss is defined as:

\begin{align*}
H(y,p) = - \sum_{i} y_{i} \log p_{i}
\end{align*}

And the mean-squared error is defined as:

\begin{align*}
MSE(y, p) = \frac{1}{N} \sum_{0}^{N} \left ( y-p \right )^{2}
\end{align*}

Our proposed loss for each gaze angle is a linear combination of the mean-squared error and cross-entropy losses, which is defined as:

\begin{align*}
CLS(y, p) = H(y,p)+ \beta \cdot MSE(y, p)
\end{align*}

Where $CLS$ is the combined loss, $p$ is the predicted values, $y$ is the ground-truth values and $\beta$ is the regression coefficient. We change the weight of the mean-squared loss during the experiments in Section \ref{sec:experiments} to obtain the best gaze performance. 

To the best of our knowledge, all related works which estimated gaze using CNN-based methods do not consider the combined classification and regression loss in their techniques.
\subsection{L2CS-Net Architecture}
We propose a simple network architecture (L2CS-Net) based upon the proposed classification and regression losses. It takes face images as input and feeds them to ResNet-50 as a backbone to extract spatial gaze features from images. In contrast to most previous work that regresses the two gaze angles together in one fully-connected layer, we predict each angle separately using two fully-connected layers. These two fully-connected layers share the same convolution layers in the backbone. Also, we use two loss functions, one for each gaze angle (yaw, pitch). Using this approach will improve network learning, as it has two signals that backpropagate through the network.

For each output from the fully-connected layer, we first use a softmax layer to convert the network output logits into a probability distribution. Then, a cross-entropy loss is applied to calculate the bin classification loss between output probabilities and target bin labels. Next, we calculate the expectation of the probability distribution to get fine-grained gaze predictions. Finally, we calculate the mean square error for this prediction and add it to the classification loss. The detailed architecture of L2CS-Net is shown in Figure \ref{fig:concept}.

\subsection{Datasets}

With the development of appearance-based gaze estimation methods, 
large-scale datasets have been proposed to improve gaze performance. These datasets were collected with different procedures, varying from laboratory-constrained settings to unconstrained indoor and outdoor environments. In order to get a valuable evaluation of our network, we train and evaluate our model using two popular datasets collected with unconstrained settings: Gaze360 and MPIIGaze.

Gaze360 \cite{Gaze360} provides the widest range of 3D gaze annotations with a range of 360 degrees. It contains 238 subjects of different ages, genders, and ethnicity. Its images are captured using a Ladybug multi-camera system in different indoor and outdoor environmental settings like lighting conditions and backgrounds.

MPIIGaze \cite{Mpiigaze1} provides 213.659 images from 15 subjects captured during their daily routine over several months. Consequently, it contains images with diverse backgrounds, time, and lighting that make it suitable for unconstrained gaze estimation. It was collected using software that asks the participants to look at randomly moving dots on their laptops.

\section{Experiments}
\label{sec:experiments}
\subsection{Data preprocessing}
We follow the same procedures in \cite{Mpiigaze1} to normalize images in the two datasets. In summary, this process applies rotation and translation to the virtual camera to remove the head's roll angle and keep the same distance between the virtual camera and a reference point (the center of the face). Furthermore, we split up the continuous gaze target in each dataset (pitch and yaw angles) into bins with binary labels for classification based on the range of the gaze annotations. As a result, both datasets have two different target annotations: continuous and binned labels make them suitable for our combined regression and classification losses. Furthermore, we change the regression coefficient in the combined loss function during the experiments to obtain the best gaze performance.

\subsection{Training and results}
We use an ImageNet-pretrained ResNet-50 as the backbone network. Our proposed network (L2CS-Net) was trained in PyTorch framework using Adam optimizer with a learning rate of 0.00001. We train our proposed network for 50 epochs using a batch size of 16. We evaluate our proposed network on MPIIGaze and Gaze360 datasets. We change the regression coefficient during the experiments and compare the output performance with the state-of-the-art gaze estimation methods. We utilize gaze angular error $(\degree)$ as the evaluation metric following most gaze estimation methods. Assuming the ground-truth gaze direction is $\mathrm{g}$  $\epsilon$ $\mathbb{R}^{3}$  and the predicted gaze vector is $\mathrm{\hat{g}}$  $\epsilon$ $\mathbb{R}^{3}$ , the gaze angular error  $(\degree)$ can be computed as:

\begin{align*}
\mathcal{L}_{angular} = \frac{\mathrm{g}\cdot \mathrm{\hat{g}}}{\left \|\mathrm{g}\right \|  \left \| \mathrm{\hat{g}} \right \|}
\end{align*}

\begin{figure}[t]
    \centering
\begin{tikzpicture}
\begin{axis}[
	ylabel={Mean Angular Error (degrees)},
	ylabel style=
    {
      yshift=-7mm, 
    },
    xmin=0,
    xmax=16.6,
    ymin=2,
    ymax=7,
	xtick = {1,2,3,4,5,6,7,8,9,10,11,12,13,14,15,16},
    xticklabels = {p00,p01,p02,p03,p04,p05,p06,p07,p08,p09,p10,p11,p12,p13,p14,Avg},
    axis on top,
    ybar=.5pt,
    x=13,
    height=7.5cm,
    my xbar legend,
    bar width=3.7,
    xtick=data,
    xtick pos=left,
    ytick style={draw=none},
	nodes near coords align={vertical},
	ymajorgrids,
    x tick label style={rotate=45,anchor=east},
    legend style={
      font=\footnotesize,legend columns=1,
                    at={(0,1.0)},anchor=north west,
                    /tikz/every even column/.append style={column sep=0.1cm}
                        },
]

\addplot[ForestGreen,fill=ForestGreen]  coordinates {

(1,2.38)
(2,2.96)
(3,3.78)
(4,3.21)
(5,2.72)
(6,4.73)
(7,3.58)
(8,4.07)
(9,5.17)
(10,3.47)
(11,4.39)
(12,6.74)
(13,3.39)
(14,4.17)
(15,4.32)
(16,3.92)
};

\addplot[YellowOrange,fill=YellowOrange]  coordinates {

(1,2.57)
(2,3.76)
(3,5.65)
(4,2.79)
(5,2.7)
(6,6.05)
(7,3.5)
(8,4.75)
(9,5.2)
(10,4.47)
(11,5.26)
(12,3.59)
(13,3.78)
(14,5.31)
(15,6.67)
(16,4.4)

};

\legend{L2CS-Net,FARE-Net}
        
\end{axis}
\end{tikzpicture}
\caption{Comparison of subject gaze accuracy between our proposed model and FARE-Net \cite{farenet} on MPIIGaze dataset.}
\label{fig:4}
\end{figure}
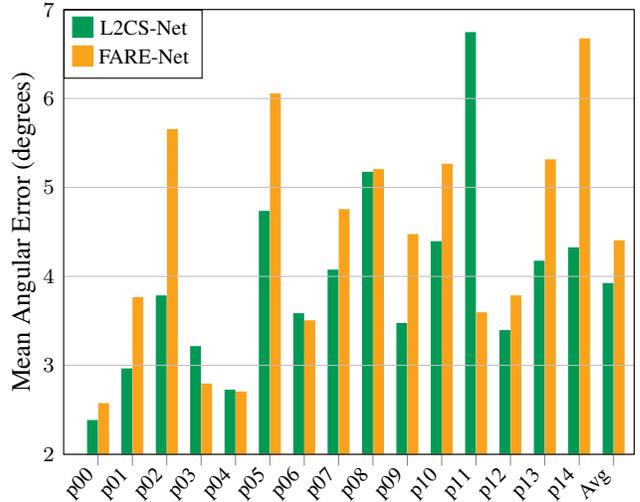

We adapt leave-one-subject-out cross-validation on MPIIGaze dataset as used in the related works \cite{Mpiigaze1,ca-net,cvpr2021}. In order to obtain the best performance, we train L2CS-Net on the MPIIGaze dataset with different regression coefficients ($\beta$) of 1 and 2. Table. \ref{tab:my-table} shows the comparison of mean angular error between our proposed model and state-of-the-art methods on MPIIGaze dataset. Our proposed L2CS-Net ($\beta=1$) achieved state-of-the-art gaze performance with 3.92\degree mean angular error. Furthermore, we present the gaze accuracy of each subject of the MPIIGaze dataset and compare it with FARE-Net \cite{farenet} as they presented the subject-wise gaze accuracy. Out of 15 subjects, our proposed method achieves better gaze accuracy for 11 subjects, as shown in Fig \ref{fig:4}.

\begin{table}[]
\resizebox*{\linewidth}{!}{%
\begin{tabular}{|c|c|}
\hline
\textbf{Methods}                                                               & \textbf{\begin{tabular}[c]{@{}c@{}}MPIIFaceGaze\end{tabular}} \\ \hline
iTracker (AlexNet) \cite{itracker}              & 5.6\degree  \\
MeNets \cite{MeNets}                            & 4.9\degree  \\
FullFace  (Spatial weights CNN)\cite{Mpiigaze1} & 4.8\degree  \\
Dilated-Net \cite{dilated}                      & 4.8\degree  \\
RT-Gene(1 model) \cite{rtgene}                  & 4.8\degree  \\
GEDDNet \cite{GEDDNet}                          & 4.5\degree  \\
RT-Gene(4 ensemble) \cite{rtgene}               & 4.3\degree  \\
Bayesian Approach \cite{BayesianApproach}       & 4.3\degree  \\
FAR-Net \cite{farenet}                          & 4.3\degree  \\
CA-Net \cite{ca-net}                            & 4.1\degree  \\
AGE-Net \cite{cvpr2021}                         & 4.09\degree \\ \hline
\textbf{\begin{tabular}[c]{@{}c@{}}L2CS-Net $(\beta=1)$\\ L2CS-Net $(\beta=2)$\end{tabular}} & \textbf{\begin{tabular}[c]{@{}c@{}}3.96\degree\\ 3.92\degree\end{tabular}}                     \\ \hline
\end{tabular}%
}
\caption{Comparison of mean angular error between our proposed model and SOTA methods on MPIIGaze dataset}
\label{tab:my-table}
\end{table}

\begin{table}[t]
\resizebox{\linewidth}{!}{%
\begin{tabular}{|c|c|c|}
\hline
\textbf{Methods} & \textbf{Front 180\degree} & \textbf{Front Facing} \\ \hline
FullFace   & 14.99\degree & N/A \\
Dilated-Net  & 13.73\degree & N/A \\ 
RT-Gene (4 ensemble)  & 12.26\degree & N/A \\ 
CA-Net  & 12.26\degree & N/A \\
Gaze360 (LSTM) \cite{Gaze360} & 11.4\degree & 11.1\degree \\ \hline
\textbf{\begin{tabular}[c]{@{}c@{}}L2CS-Net  $(\beta =2)$\\ L2CS-Net  $(\beta =1)$\end{tabular}} & \textbf{\begin{tabular}[c]{@{}c@{}}10.54\degree\\ 10.41\degree\end{tabular}} & \textbf{\begin{tabular}[c]{@{}c@{}}9.13\degree\\ 9.02\degree\end{tabular}} \\ \hline
\end{tabular}%
}
\caption{Comparison of mean angular error between our proposed model and SOTA methods on Gaze360 dataset}
\label{tab:my-table1}
\end{table}

Kellnhofer et al. \cite{Gaze360} divided the Gaze360 dataset into train-val-test sets and presented three evaluation scopes based on the range of gaze angles: all 360\degree, front 180\degree, and front-facing (within 20\degree). We follow the same evaluation criteria in \cite{Gaze360}, but only with the front 180\degree and front-facing for a fair comparison with all related methods that are trained and evaluated on datasets within 180\degree range. We trained L2CS-Net on Gaze360 dataset with different regression coefficients of 1 and 2. Table. \ref{tab:my-table1} shows the comparison of mean angular error between our proposed model and State-of-the-art methods on Gaze360 dataset. We used the results from \cite{survey} as they implement typical gaze estimation methods on the Gaze360 dataset. Our proposed L2CS-Net ($\beta=1$) achieves state-of-the-art gaze performance with 10.41\degree mean angular error on front 180\degree and 9.02\degree on front facing.

\section{Conclusion}
\label{sec:conclusion}
In this paper, we present a robust CNN-based model (L2CS-Net) for predicting 3D gaze directions in unconstrained environments. We propose to predict each gaze angle individually with two fully-connected layers and a ResNet-50 backbone. In order to improve the network learning, we used two separate loss functions for each gaze angle, each of them is a linear combination of regression and classification losses. Further, we use a reliable softmax layer to predict gaze bins. Furthermore, we changed the regression coefficient during the experiments to obtain the best gaze performance. To show the robustness of our model, we validate our network using two of the most unconstrained gaze datasets: MPIIGaze and Gaze360, and we followed the same evaluation criteria used in each dataset. Our model achieved state-of-the-art gaze accuracy with the lowest angular error in both datasets.

\bibliographystyle{IEEEbib}
\bibliography{ref}

\begin{thebibliography}{10}

\bibitem{hempel2020slam}
Thorsten Hempel and Ayoub Al-Hamadi,
\newblock ``Slam-based multistate tracking system for mobile human-robot
  interaction,''
\newblock in {\em International Conference on Image Analysis and Recognition}.
  Springer, 2020, pp. 368--376.

\bibitem{strazdas2022robot}
Dominykas Strazdas, Jan Hintz, Aly Khalifa, Ahmed~A Abdelrahman, Thorsten
  Hempel, and Ayoub Al-Hamadi,
\newblock ``Robot system assistant (rosa): Towards intuitive multi-modal and
  multi-device human-robot interaction,''
\newblock {\em Sensors}, vol. 22, no. 3, pp. 923, 2022.

\bibitem{li2012vision}
Liyuan Li, Xinguo Yu, Jun Li, Gang Wang, Ji-Yu Shi, Yeow~Kee Tan, and Haizhou
  Li,
\newblock ``Vision-based attention estimation and selection for social robot to
  perform natural interaction in the open world,''
\newblock in {\em 2012 7th ACM/IEEE International Conference on Human-Robot
  Interaction (HRI)}. IEEE, 2012, pp. 183--184.

\bibitem{patney2016towards}
Anjul Patney, Marco Salvi, Joohwan Kim, Anton Kaplanyan, Chris Wyman, Nir
  Benty, David Luebke, and Aaron Lefohn,
\newblock ``Towards foveated rendering for gaze-tracked virtual reality,''
\newblock {\em ACM Transactions on Graphics (TOG)}, vol. 35, no. 6, pp. 1--12,
  2016.

\bibitem{cvpr2021}
Pradipta Biswas et~al.,
\newblock ``Appearance-based gaze estimation using attention and difference
  mechanism,''
\newblock in {\em Proceedings of the IEEE/CVF Conference on Computer Vision and
  Pattern Recognition}, 2021, pp. 3143--3152.

\bibitem{ca-net}
Yihua Cheng, Shiyao Huang, Fei Wang, Chen Qian, and Feng Lu,
\newblock ``A coarse-to-fine adaptive network for appearance-based gaze
  estimation,''
\newblock in {\em Proceedings of the AAAI Conference on Artificial
  Intelligence}, 2020, vol.~34, pp. 10623--10630.

\bibitem{dilated}
Zhaokang Chen and Bertram~E Shi,
\newblock ``Appearance-based gaze estimation using dilated-convolutions,''
\newblock in {\em Asian Conference on Computer Vision}. Springer, 2018, pp.
  309--324.

\bibitem{liu2020gaze}
Song Liu, Danping Liu, and Haiyang Wu,
\newblock ``Gaze estimation with multi-scale channel and spatial attention,''
\newblock in {\em Proceedings of the 2020 9th International Conference on
  Computing and Pattern Recognition}, 2020, pp. 303--309.

\bibitem{Gaze360}
Petr Kellnhofer, Adria Recasens, Simon Stent, Wojciech Matusik, and Antonio
  Torralba,
\newblock ``Gaze360: Physically unconstrained gaze estimation in the wild,''
\newblock in {\em Proceedings of the IEEE/CVF International Conference on
  Computer Vision}, 2019, pp. 6912--6921.

\bibitem{Mpiigaze}
Xucong Zhang, Yusuke Sugano, Mario Fritz, and Andreas Bulling,
\newblock ``Mpiigaze: Real-world dataset and deep appearance-based gaze
  estimation,''
\newblock {\em IEEE transactions on pattern analysis and machine intelligence},
  vol. 41, no. 1, pp. 162--175, 2017.

\bibitem{rtgene}
Tobias Fischer, Hyung~Jin Chang, and Yiannis Demiris,
\newblock ``Rt-gene: Real-time eye gaze estimation in natural environments,''
\newblock in {\em Proceedings of the European Conference on Computer Vision
  (ECCV)}, 2018, pp. 334--352.

\bibitem{Ethxgaze}
Xucong Zhang, Seonwook Park, Thabo Beeler, Derek Bradley, Siyu Tang, and Otmar
  Hilliges,
\newblock ``Eth-xgaze: A large scale dataset for gaze estimation under extreme
  head pose and gaze variation,''
\newblock in {\em European Conference on Computer Vision}. Springer, 2020, pp.
  365--381.

\bibitem{lu2014adaptive}
Feng Lu, Yusuke Sugano, Takahiro Okabe, and Yoichi Sato,
\newblock ``Adaptive linear regression for appearance-based gaze estimation,''
\newblock {\em IEEE transactions on pattern analysis and machine intelligence},
  vol. 36, no. 10, pp. 2033--2046, 2014.

\bibitem{williams2006sparse}
Oliver Williams, Andrew Blake, and Roberto Cipolla,
\newblock ``Sparse and semi-supervised visual mapping with the s\^{} 3gp,''
\newblock in {\em 2006 IEEE Computer Society Conference on Computer Vision and
  Pattern Recognition (CVPR'06)}. IEEE, 2006, vol.~1, pp. 230--237.

\bibitem{Mpiigaze1}
Xucong Zhang, Yusuke Sugano, Mario Fritz, and Andreas Bulling,
\newblock ``It’s written all over your face: Full-face appearance-based gaze
  estimation,''
\newblock in {\em Computer Vision and Pattern Recognition Workshops (CVPRW),
  2017 IEEE Conference on}. IEEE, 2017, pp. 2299--2308.

\bibitem{itracker}
Kyle Krafka, Aditya Khosla, Petr Kellnhofer, Harini Kannan, Suchendra
  Bhandarkar, Wojciech Matusik, and Antonio Torralba,
\newblock ``Eye tracking for everyone,''
\newblock in {\em Proceedings of the IEEE conference on computer vision and
  pattern recognition}, 2016, pp. 2176--2184.

\bibitem{MeNets}
Yunyang Xiong, Hyunwoo~J Kim, and Vikas Singh,
\newblock ``Mixed effects neural networks (menets) with applications to gaze
  estimation,''
\newblock in {\em Proceedings of the IEEE/CVF Conference on Computer Vision and
  Pattern Recognition}, 2019, pp. 7743--7752.

\bibitem{GEDDNet}
Zhaokang Chen and Bertram~E Shi,
\newblock ``Geddnet: A network for gaze estimation with dilation and
  decomposition,''
\newblock {\em arXiv preprint arXiv:2001.09284}, 2020.

\bibitem{farenet}
Yihua Cheng, Xucong Zhang, Feng Lu, and Yoichi Sato,
\newblock ``Gaze estimation by exploring two-eye asymmetry,''
\newblock {\em IEEE Transactions on Image Processing}, vol. 29, pp. 5259--5272,
  2020.

\bibitem{BayesianApproach}
Kang Wang, Rui Zhao, Hui Su, and Qiang Ji,
\newblock ``Generalizing eye tracking with bayesian adversarial learning,''
\newblock in {\em Proceedings of the IEEE/CVF Conference on Computer Vision and
  Pattern Recognition}, 2019, pp. 11907--11916.

\bibitem{survey}
Yihua Cheng, Haofei Wang, Yiwei Bao, and Feng Lu,
\newblock ``Appearance-based gaze estimation with deep learning: A review and
  benchmark,''
\newblock {\em arXiv preprint arXiv:2104.12668}, 2021.

\end{thebibliography}



\end{document}